\documentclass[12pt]{article}

\usepackage{newtxtext,newtxmath}
\usepackage{graphicx}
\usepackage[letterpaper,margin=1in]{geometry}
\usepackage{booktabs}
\renewenvironment{abstract}
	{\quotation}
	{\endquotation}
\date{}

\makeatletter
\renewcommand{\fnum@figure}{\textbf{Figure \thefigure}}
\renewcommand{\fnum@table}{\textbf{Table \thetable}}
\makeatother

\usepackage{scicite}
\usepackage{url}
\usepackage{listings}
\usepackage{xcolor}
\definecolor{codegray}{RGB}{248,248,248}
\definecolor{codeblue}{RGB}{35,85,140}
\definecolor{codegreen}{RGB}{40,120,70}
\definecolor{codeorange}{RGB}{170,90,45}
\definecolor{codepurple}{RGB}{120,70,150}
\lstdefinestyle{nneql}{
    backgroundcolor=\color{codegray},
    basicstyle=\ttfamily\small,
    frame=single,
    rulecolor=\color{black!15},
    breaklines=true,
    showstringspaces=false,
    columns=fullflexible,
    xleftmargin=0.4em,
    framexleftmargin=0.4em,
    aboveskip=0.8em,
    belowskip=0.8em
}
\lstdefinelanguage{yaml}{
    comment=[l]{\#},
    commentstyle=\color{codegreen}\itshape,
    string=[b]',
    morestring=[b]",
    stringstyle=\color{codeorange},
    keywords={true,false,null,yes,no},
    keywordstyle=\color{codepurple}\bfseries,
    sensitive=false
}

\def\scititle{
	Interpretable neural equation learning for high-dimensional materials data
}
\title{\bfseries \boldmath \scititle}

\author{
	Philip Jacobson$^{1\ast}$,
	Prasad Iyer$^{2}$,
	Remi Dingreville$^{2}$,
    Saaketh Desai$^{2\ast}$\and
	\small$^{1}$Sandia National Laboratories, Livermore, CA, USA.\and
	\small$^{2}$Center for Integrated Nanotechnologies, Sandia National Laboratories, Albuquerque , NM, USA.\and
	\small$^\ast$Corresponding author. Email: saadesa@sandia.gov.
}

\begin{document} 

\maketitle

\begin{abstract} \bfseries \boldmath
Equation learning seeks to discover compact, interpretable equations from data using various machine-learning techniques.
However, few workflows discover and interpret learned equations in sparse, high-dimensional materials datasets.
We present nn-EQL, an equation-learning framework that generates candidate equations through iterative training, pruning, and retraining of a neural network with customizable, user-specified neuron-level activation functions.
Candidate equations are ranked using a minimum description length (MDL)-inspired score, balancing error and sparsity, and diagnosed to assess term-wise contributions, input-scaled sensitivities, and uncertainties.
We demonstrate nn-EQL on standard AI-Feynman benchmarks as well as multiscale high-dimensional materials datasets such as finite-element simulations of heterogeneous materials, and high-throughput experimental thin-film deposition data. 
nn-EQL integrates with tabular data and high-dimensional data via standard encoder models, learning equations and providing insight into learned terms.
Looking forward, integrating physically interpretable representations, multimodal and multi-fidelity data, and adaptive data acquisition could enable equation learning to formulate and test mechanistic hypotheses. 
\end{abstract}

\section*{Introduction}
\subsection*{Learning equations from scientific data}

Data-driven techniques, particularly deep-learning approaches, have shown excellent predictive capabilities when working with materials data \cite{choudhary2022recent, merchant2023scaling, lysogorskiy2026graph, oommen2024rethinking, lu2020extraction}, focusing on ingesting large datasets and learning correlations to rapidly predict properties \cite{desai2024navigating, jha2018elemnet} without performing expensive simulations or experiments.
Other techniques such as Bayesian optimization and data-driven design of experiments focus even more on sample efficiency, minimizing the number of experiments or simulations needed to maximize or minimize a material property \cite{kusne2020fly, wakabayashi2019machine}.
While these techniques focus on predictive accuracy, or sample efficiency in directing experiments, scientific understanding requires compact and interpretable relationships among variables.
Equations are a traditional form of interpreting relationships among variables, and can expose limits of current theory, rank competing mechanisms, and propose a hierarchy for candidate descriptors affecting the property of interest \cite{wang_exploring_2024, krenn_scientific_2022}.

As a specific instance of this distinction, fatigue of metallic alloys is caused by a number of hierarchical, co-acting mechanisms such as grain boundary sliding, slip band formation, dislocation pile-up \cite{boyce2026materials, barr2023autonomous}. 
Data-driven deep learning techniques can predict fatigue life from data, but lack interpretability and do not have the ability to rank the mechanisms listed above.
In contrast, conventional theories such as Paris law list explicit equations relating crack growth rates to stress intensity factors, which in turn relate to material specific mechanisms \cite{chowdhury2016mechanisms}.
In this exemplar, specific equations relating crack growth to material microstructure descriptors would combine the best of data-driven approaches and traditional equation learning, opening up a frontier of mechanisms which can be investigated with subsequent experiments or multiscale simulations.
 Similarly, the hardness of nanocrystalline thin-film alloys is a result of Hall-Petch strengthening, residual stress, and solid-solution strengthening, all reflected in microstructural fingerprints such as X-ray diffractograms.
 Deep learning can predict hardness from diffractograms \cite{desai2024navigating}, but alone cannot identify microstructural features that matter. An equation learned directly from the diffractograms would isolate such features, forming a hypothesis that targeted experiments can test.

Learning equations from data, i.e., equation learning, or symbolic regression, is more than curve fitting, and can outperform traditional machine learning methods for small datasets \cite{wilstrup_symbolic_2021}.
While curve fitting assumes a fixed functional form and finds coefficients associated with the functional form, equation learning finds a set of functional forms and coefficients from the vast space of possible combinations.
In other words, curve fitting is symbolic regression with inductive bias from the researcher assuming a functional form apriori, based on prior knowledge and literature.

The fundamental challenge in equation learning for scientific discovery is limited datasets when exploring novel scientific phenomena, which results in multiple \textbf{generated} functional forms fitting the dataset well, with each functional form representing different hypotheses about the mechanisms or variable interactions underlying the observed response.
This model \textbf{selection} problem is even more acute in material science, where material characterization data is often sparse, noisy, and sampled over limited spatial/temporal domains.

Symbolic regression techniques often address this challenge by focusing on providing a set of candidate equations, ranking them in some form to indicate data-based fit, and suitability to the domain \cite{schmidt_distilling_2009, udrescu_ai_2020-1}.
One common form of ranking is in the form of a pareto-front comparing accuracy and complexity, invoking the Occam's razor principle: given equal accuracy, the explanation (equation) with lower complexity is more likely to represent the underlying mechanism.
However, this \textbf{diagnosis} does not resolve the scientific question, since the true equation is unknown in many scientific applications, and several candidate expressions may provide comparable and accurate descriptions of the data while implying different variable interactions or mechanisms, i.e., different \textbf{interpretations}.

A related and increasingly important challenge in the modern area of data-driven scientific discovery is dimensionality, i.e., choosing the variables the equations should be written in.
Most symbolic regression techniques are designed to provide a closed-form equation from a modest number of scalar candidate variables that a researcher has already identified as physically relevant, and the size of the search space over functional forms grows rapidly with the number of candidate variables considered.
Many materials datasets, however, are not naturally described by a handful of scalar descriptors: microstructure images, diffraction patterns, and spectroscopy measurements may contain hundreds to thousands of raw input dimensions, with no obvious reduction to a few hand-picked variables. Extending equation learning to operate on such high-dimensional, structured data directly remains largely unaddressed by existing symbolic regression methods \cite{tsoi2025symbolnet, reinbold2021robust} and is a central capability we highlight in this work.

We therefore need equation learning workflows that can \textbf{generate} candidate equations, \textbf{select} among competing hypotheses, \textbf{diagnose} the ambiguity and robustness of terms, and help scientists \textbf{interpret} and \textbf{refine} them through additional analysis, especially for high-dimensional data, where the variables of the equation should be traceable to the raw data and comparable with established physical descriptors.
An equation that meets these requirements can be read as a set of testable hypotheses about the system.
In what follows, we first cover the landscape of equation learning, covering various algorithms and approaches to learn equations, along with datasets to benchmark equation learners.
We then describe our neural network based equation learner and our approach to generate, select, diagnose, and interpret equations learned from data.
We then show benchmarking results and accompanying interpretability analyses on known datasets, following which we apply our workflow to a materials simulation datasets at the atomistic and the mesoscale, and a high-throughput combinatorial experimental dataset, finding novel equations for materials phenomena at various length scales, and interpreting the learned equations.
Our neural network based approach makes our equation learning workflow amenable to high-dimensional datasets, a typical limitation of genetic programming approaches, while retaining interpretability.

\subsection*{The landscape of equation learning}

\textbf{Algorithms to generate and select equations from data:}
Much of the literature on equation learning has focused on symbolic regression, i.e., the task of developing algorithms to \textbf{generate} and \textbf{select} candidate equations directly from data.
The foundations of symbolic regression were set in part by Koza \cite{koza_genetic_1992}, coining the term genetic programming for a procedure that involved genetic operations (crossovers, mutations) to write programs.
This general genetic approach was further developed by Schmidt and Lipson \cite{schmidt_distilling_2009} in their software Eureqa, applying genetic approaches to \textbf{generate} equations directly from experimental data.
This work also popularized a fundamental \textbf{selection} tradeoff in symbolic regression between accuracy and complexity, invoking the Occam's razor to identify a Pareto front that helped select the simplest equations with desired accuracy.

Building on this work, a variety of methods have proposed models to \textbf{generate} and \textbf{select} equations directly from data, ranging from genetic algorithms \cite{cranmer_discovering_2020}, shallow feedforward neural networks \cite{martius_extrapolation_2016, sahoo_learning_2018, desai_parsimonious_2021}, Monte Carlo tree-based approaches \cite{kamienny_deep_2023}, and sparse regression techniques \cite{brunton_discovering_2016} (applied primarily to differential equations).
Purely deep learning based approaches have also gained popularity in recent times, focusing on generative approaches for terms in an equation using VAEs \cite{meznar_efficient_2023, holt_deep_2023}, reinforcement learning \cite{mundhenk_symbolic_2021}, transformer based techniques \cite{kamienny_end--end_2022, shojaee_transformer-based_2023}, and even LLM-based techniques \cite{shojaee_llm-sr_2025}.
Specialized networks such as Kolmogorov-Arnold networks \cite{liu_kan_2025, he_mlp-kan_2024} have also gained popularity in recent times \cite{buhler_kan-sr_2025}.

\textbf{Equation diagnosis, interpretation, and refinement:}
Comparatively less attention has been paid to what happens after multiple plausible equations have been generated and selected, particularly how to \textbf{diagnose} ambiguity among competing expressions and \textbf{interpret} their scientific implications.
Several notable approaches already broaden equation learning beyond unconstrained symbolic search.
AI-Feynman \cite{udrescu_ai_2020} includes a scientific structure in the equation generation process with dimensionality analysis, variable separation, and polynomial preliminary fitting.
AI-Descartes \cite{cornelio_combining_2023} goes a step further and combines symbolic regression with axiomatic reasoning to provide additional context for the learning of equations, beyond the provided data set.
These approaches demonstrate that information outside of predictive fit and equation complexity can play an important role in equation learning.
Finally, learned equations can be \textbf{refined} using uncertainty or disagreement among learned equations to guide the acquisition of additional data.
Recent approaches that combine symbolic regression with active learning demonstrate this idea by selecting new measurements or trajectories that are informative to distinguish or improve candidate equations \cite{ansari2022iterative, medina2023active, bomarito2025bayesian}.

\textbf{Evaluating equation learners:}
Given the wide variety of equation learning techniques, benchmark datasets have become essential for evaluating how well various techniques \textbf{generate} and \textbf{select} candidate equations.
La Cava et al. introduced SRBench \cite{la_cava_contemporary_2021}, as an empirical benchmark with known equations to systematically compare symbolic regression methods across a broad collection of problems.
In conjunction with synthetic benchmarks such as AI-Feynman \cite{udrescu_ai_2020}, and the more recent LSR-Synth \cite{shojaee_llm-sr_2025}, discovered equations can be evaluated for predictive accuracy, equation complexity, and recovery of known functional forms.

More recent benchmarking efforts also show the need to \textbf{diagnose} equation-learning behavior beyond a single accuracy metric.
Benchmarks such as SRSD \cite{matsubara_rethinking_2024},SRBench++\cite{de2024srbench++}, and recent calls for next-generation symbolic-regression benchmarks\cite{imai2025call} emphasize that performance depends on factors such as data sampling, noise, structural similarity, model complexity, and the inductive biases of the equation learner.
An equation can be accurate but unnecessarily complex, simple but structurally incorrect, or successful in one data regime but unstable in another~\cite{de2024srbench++, voigt2025analyzing}.

However, benchmarking primarily evaluates the capability of an equation learning technique across controlled problems as a success or a failure.
Scientific use of equation learners requires an additional level of evidence about the particular equations learned from a particular dataset, e.g. how should the learned equation be \textbf{interpreted}, whether competing expressions represent meaningfully different hypotheses, and what additional data would \textbf{refine} or distinguish those hypotheses.
In this work, we first use benchmark problems to establish nn-EQL as a capable method for \textbf{generating} and \textbf{selecting} equations, and then use it as a vehicle to examine the broader tasks of \textbf{diagnosing}, \textbf{interpreting}, and \textbf{refining} learned equations.

\section*{Results}
\subsection*{nn-EQL: A neural network based equation learner}

\begin{figure}[!ht]
\centering
  \includegraphics[width=\columnwidth]{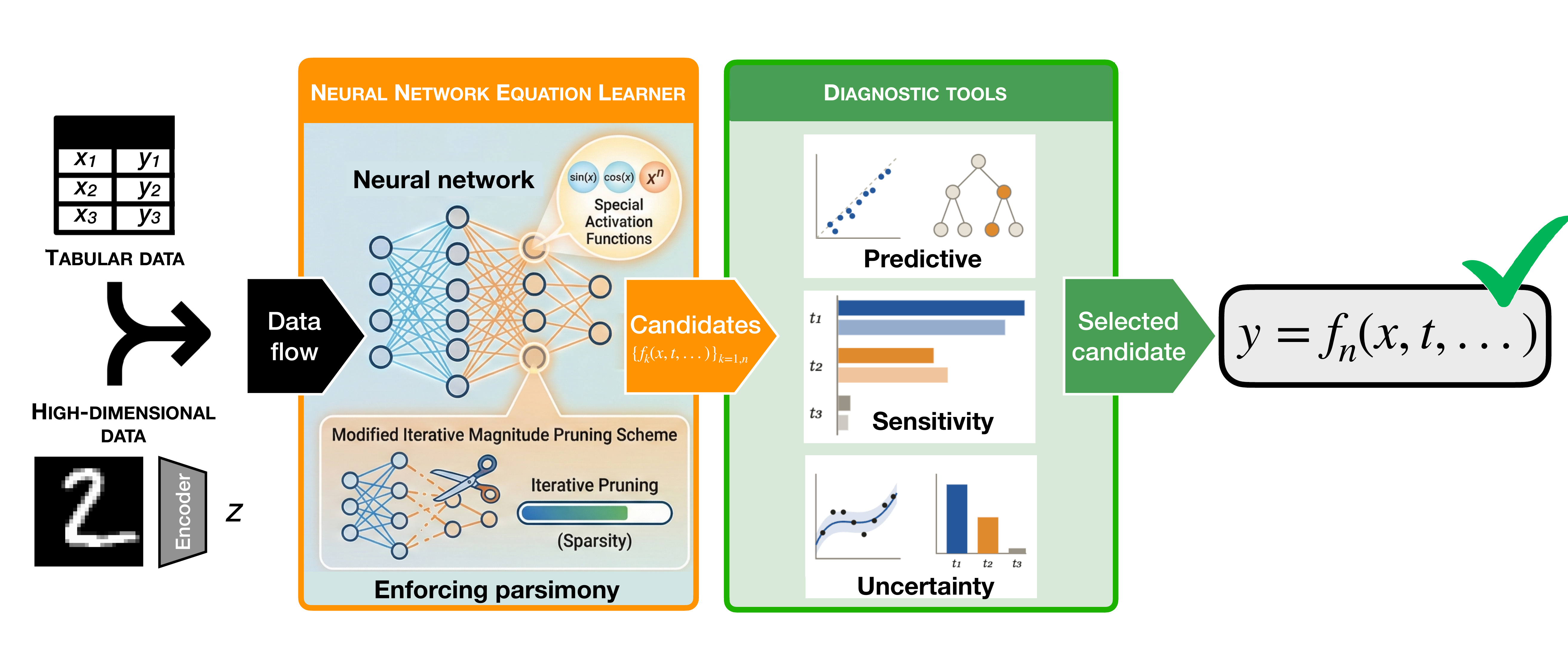}
  \caption{{\textbf{nn-EQL workflow.} A set of  candidate equations are generated by iteratively pruning a large neural network with customized activation functions. A set of diagnostic tools are then used for assessing the quality of the prediction, including term wise contributions, sensitivities, and uncertainty. Those metrics are ultimately used to determine the equation that best fits the data.}
  }
  \label{fig:eql_workflow}
\end{figure}

As illustrated in Fig\@.~\ref{fig:eql_workflow}, we use a neural network equation learner (nn-EQL) as a concrete platform for illustrating the broader equation learning workflow developed in this work.
nn-EQL builds on our prior work \cite{desai_parsimonious_2021, desai2026learning} to \textbf{generate} candidate equations by iteratively pruning a large neural network with customized activation functions.
Model selection criteria such as minimum description length provide mechanisms to \textbf{select} compact expressions from equations generated via pruning.
A neural network based representation of equations additionally provides us quantities that can be used to \textbf{diagnose} ambiguity and \textbf{interpret} learned expressions. 
Finally, nn-EQL can be coupled to active learning strategies \cite{desai2025autoscilab} to \textbf{refine} candidate equations/hypotheses with additional data.

\textbf{Model architecture:}
Our equation learner model, nn-EQL, is a neural network with customized activation functions per neuron, providing the ability to combine and compose functional forms via connections between neurons in different layers.
This approach conceptually builds on EQL \cite{martius_extrapolation_2016} to generate candidate equations.

Formally, let $\mathbf{x} \in {\mathbb{R}}^{d}$ denote the inputs to nn-EQL and let the activation at layer $\ell$ be denoted by $h^{(\ell)}$, such that $h^{(0)} = \mathbf{x}$.
For a hidden layer, linear combinations of the preceding activations are first computed as
\begin{equation}
z_j^{(\ell)}=\sum_i W_{ji}^{(\ell)}h_i^{(\ell)}; \quad h_j^{(\ell+1)}=\phi_j^{(\ell)} \left(z_j^{(\ell)}\right).
\end{equation}
as is the case in a standard neural network, with the only difference being that the functions $\phi_j^{(\ell)}$ are selected from a user-defined operator library containing functions such as for instance
\begin{equation}
\phi \in \{x, x^2, x^3, \sqrt x, \sin (x), \cos (x), \exp (x), \log (x)\},
\end{equation}
as opposed to a fixed tanh or ReLU activation function applied to all neurons in a layer in a conventional neural network.
Note that in addition to the above unary operators, we also provide a binary multiplication operator that combines two activations $z_j^{(\ell)}$ and $z_{j+1}^{(\ell)}$ .
The operator library can therefore be adapted to the scientific problem and to functional forms suggested by prior knowledge.
In the implementation used here, two operator layers are followed by a linear output layer,
\begin{equation}
\hat{y}=\sum_j W_j^{(L)}h_j^{(L)}+b^{(L)}.
\end{equation}
Because every retained operation has an explicit analytic form, the trained network can be composed directly into a symbolic expression.


We follow a three step iterative train-prune-retrain approach to \textbf{generate} candidate equations.

\textbf{(i) Fitting an overparameterized network: } Given a dataset, our workflow involves first training a dense nn-EQL network, characterized by every neuron being connected to every other neuron, as is the convention in standard feedforward neural networks.
This dense network is trained to minimize the mean squared error (MSE), with an Adam optimizer \cite{kingma2014adam}, although our PyTorch implementation supports other objective functions and optimizers.
We begin from an intentionally overparameterized network to ensure sufficient capacity to fit the dataset at hand.
The universal approximation theorem establishes that sufficiently expressive feedforward neural networks can approximate continuous functions to arbitrary accuracy over compact domains under appropriate assumptions \cite{hornik1989multilayer, cybenko1989approximation}, motivating our use of an overparameterized network as the starting point for equation generation.
The operator library can additionally contain conventional nonlinear activations such as $\tanh$, providing a flexible neural network representation when a compact combination of the other analytic operators cannot adequately fit the data.
This initial step generates an equation:
\begin{equation}
\hat{y} = f_0(\textbf{x}) = \rm nn-EQL(\textbf{x})
\end{equation}

\textbf{(ii) Prune connections in the network: } Starting from the dense network, candidate equations are then generated through iterative pruning of this trained network.
Pruning is quite common in neural networks \cite{frankle2018lottery}, operating on the hypothesis that a dense network contains a sparse subnetwork that still fits a dataset, and generalizes to unseen data well.
For a connection from activation $h_i^{(\ell-1)}$ to $h_i^{(\ell)}$ with weight $W_{ji}^{(\ell)}$, we define a data-dependent contribution score as
\begin{equation}
s_{ji}^{(\ell)} = \left|W_{ji}^{(\ell)} \left\langle h_i^{(\ell)}(\mathbf{x}) \right\rangle_{\mathbf{x}\in D} \right|,
\end{equation}
where $\langle\cdot\rangle_{\mathbf{x}\in D}$ denotes an average over the training dataset.
At each pruning step, a prescribed fraction of the lowest-contribution connections is removed from all layers.
Note that this `contribution-based' pruning is a customized version of standard pruning, since conventional pruning focuses on removing weights with lowest magnitudes.
However, nn-EQL has neurons with different activation functions, making weight magnitude alone an inappropriate measure of connection importance.
For example, a small weight attached to a high magnitude activation may contribute
more to the output than a larger weight attached to an activation that is nearly
zero over the dataset.

\textbf{(iii) Retrain pruned network: } The remaining network is then retrained while the pruned connections are constrained to remain zero.
This allows the remaining weights to adjust, with the number of retraining epochs set as a hyperparameter, and smooths out the sudden degradation in fit quality \cite{desai2025self}.
This train-prune-retrain process is repeated with progressively increasing sparsity to generate a set of networks, each of which is read out as an explicit symbolic equation by composing the weights and activation functions using packages as Sympy \cite{meurer2017sympy}.

The result of the train-prune-retrain process described above is a set of candidate equations, with increasing parsimony:
\begin{equation}
\mathcal{F} = \{ f_0(\mathbf{x}), f_1(\mathbf{x}),f_2(\mathbf{x}),\ldots,f_K(\mathbf{x}) \}
\end{equation}

\textbf{Selecting candidate equations:}
The family of candidate equations $\mathcal{F}$ exhibits varying levels of predictive accuracy and complexity.
A separate model-selection step is therefore needed to rank these candidates according to their ability to explain the observed data while remaining parsimonious, as is standard practice in other equation learners \cite{cranmer_discovering_2020}.

A common framework in model selection for balancing fit and complexity is the minimum description length (MDL) principle \cite{grunwald2007minimum}, which favors models that provide a concise description of both the data and the model itself.
In symbolic regression, however, the definition of model description length depends on how an equation is represented.
For methods that express an equation as a tree, complexity can be associated naturally with quantities such as tree size, number of operators, or expression length \cite{bartlett2023exhaustive, yu2025symbolic}.
In nn-EQL, candidate equations are represented during training as sparse neural computational graphs, making a representation-independent definition of equation description length less direct.
Equivalents such as number of weights are often used as regularization tools \cite{martius_extrapolation_2016}, or explicitly in MDL-based neural network training \cite{bartlett2023exhaustive, yu2025symbolic}, but the applicability of these definitions to nn-EQL is yet to be determined.
We instead use a heuristic MDL-inspired criterion to rank candidate nn-EQL models, and provide \textbf{diagnostic} tools for each candidate equation to empower the user to better select equations.
We define
\begin{equation}
S_k = \frac{N}{2}\log\left(\mathrm{MSE}_k\right) + \lambda N_{W_k},
\end{equation}
where $N$ is the number of data points, $\mathrm{MSE}_k$ is the mean-squared error of candidate $f_k$, $N_{W_k}$ is the number of active network connections, and the hyperparameter $\lambda$ (typically between 0.25-1.0) controls the penalty assigned to model complexity.
The first term favors agreement with the observed data, while the second favors sparse network representations, making this a variant of the conventional accuracy-complexity trade-off \cite{schmidt_distilling_2009}, but with a connection to more formal MDL based definitions.

The score is evaluated across all models generated along the pruning trajectory $
\mathcal{F} = \{ f_0,f_1,\ldots,f_K \} $ providing a ranking of candidate equations across the accuracy-complexity tradeoff.
We note that while we rank candidate equations, we do not prescribe an optimal equation to the user, because the complexity term is representation dependent and serves only as a heuristic proxy for description length.
Instead, the full family of candidate equations is retained and reported, with the \textbf{selection} ranking treated as one tool to help users select equations.

\textbf{Diagnosing and interpreting candidate equations:}
Selection based on predictive accuracy and model complexity does not fully establish whether a learned equation is uniquely supported by the available data or whether its individual terms have meaningful scientific significance.
We therefore associate each equation from the step above with a set of diagnostics that characterize (i) predictive performance, (ii) structural fidelity, (iii) the role of individual terms (iv) robustness and uncertainty.
These diagnostics give users complementary evidence for evaluating each term of a candidate equation.
They fall into three categories: predictive capability, term contributions and sensitivities, and predictive uncertainty with term-wise support.

\textbf{Predictive capability:} For each candidate equation, we quantify performance on held-out test data using classical regression metrics such as  $R^2$, root-mean-squared error (RMSE), and mean absolute error (MAE).
For benchmark problems where the ground-truth equation $f^\ast$ is known, we additionally diagnose whether a learned equation has recovered the correct functional \textbf{structure}.
We test symbolic equivalence between the learned and true expressions using a normalized tree-edit distance (NED) \cite{matsubara_rethinking_2024}.
Each equation is represented as a symbolic expression tree, and we compute
\begin{equation}
D_{\mathrm{NED}}(f_k,f^\ast) = \frac{D_{\mathrm{TED}} \left(T(f_k),T(f^\ast)\right)} {|T(f^\ast)|},
\end{equation}
where $T(f)$ denotes the expression tree associated with equation $f$, $D_{\mathrm{TED}}$ is the tree-edit distance required to transform one expression tree into the other, and $|T(f^\ast)|$ is the number of nodes in the reference tree.
Numerical coefficients can be removed prior to comparison so that the diagnostic emphasizes recovery of functional structure rather than exact parameter values.

\textbf{Term contributions and sensitivities:}
Decomposing equations into additive terms,
\begin{equation}
f_k(\mathbf{x}) = \sum_{j=1}^{M} t_j(\mathbf{x}),
\end{equation}
we define the normalized contribution of term $t_j$ at each data point as
\begin{equation}
C_j(\mathbf{x}) = \frac{|t_j(\mathbf{x})|} {\sum_{m=1}^{M}|t_m(\mathbf{x})| + \epsilon},
\end{equation}
where $\epsilon$ is a small numerical constant. Averaging $C_j$ over the dataset provides a global indication of the relative magnitude of each term, while its point-wise values identify regions of the input domain where individual terms become dominant.
We further quantify the response of each term to the input variables through its gradients, via input-scaled sensitivities.
\begin{equation}
G_{jd}(\mathbf{x}) = \frac{\partial t_j(\mathbf{x})}{\partial x_d} \quad S_{jd}(\mathbf{x}) = \left|x_d \frac{\partial t_j(\mathbf{x})}{\partial x_d} \right|.
\end{equation}

\textbf{Predictive uncertainty and term-wise support: } When a separate calibration dataset is available, we quantify predictive uncertainty using split conformal prediction \cite{shafer2008tutorial}.
Conformal prediction is commonly used in black-box machine learning approaches to provide guaranteed statistical bounds on model predictions, especially in domains such as drug-discovery \cite{bosc2019large, alvarsson2021predicting}.
For a candidate equation $f_k$, absolute residuals on the calibration set are defined as
$r_n = \left|y_n-f_k(\mathbf{x_n})\right|$.
For a desired miscoverage level $\alpha$, the corresponding finite-sample conformal quantile $\hat q_{1-\alpha}$ defines the prediction interval $f_k(\mathbf{x}) \pm \hat q_{1-\alpha}$.
The width of this interval provides a direct measure of the residual uncertainty required for the candidate equation to achieve the desired empirical coverage.

We additionally diagnose whether each term is supported by the available data.
Each term $t_j$ is removed and the conformal interval is re-calibrated without refitting the remaining equation. We define
\begin{equation}
\Delta \hat q_j = \hat q_{-j}-\hat q
\end{equation}
where $\hat q$ is the conformal quantile for the complete expression and $\hat q_{-j}$ is the corresponding quantity after removing term $t_j$.
A large positive $\Delta\hat q_j$ indicates that removing the term substantially increases the uncertainty required to maintain the target coverage and therefore provides evidence that the term contributes meaningful predictive information.
Conversely, small changes in the conformal width indicate that the available data provides weaker evidence that the term is necessary.
The normalized contribution and input-scaled sensitivity measure how large a term is and how strongly it responds to its inputs. 
$\Delta\hat{q}_j$ measures whether the data needs that term.
A term with a large contribution can still have small $\Delta\hat{q}_j$, for example when another term nearly duplicates it.





\subsection*{Benchmarking on known equations}

\begin{figure}[!ht]
\centering
\includegraphics[width=0.55\columnwidth]{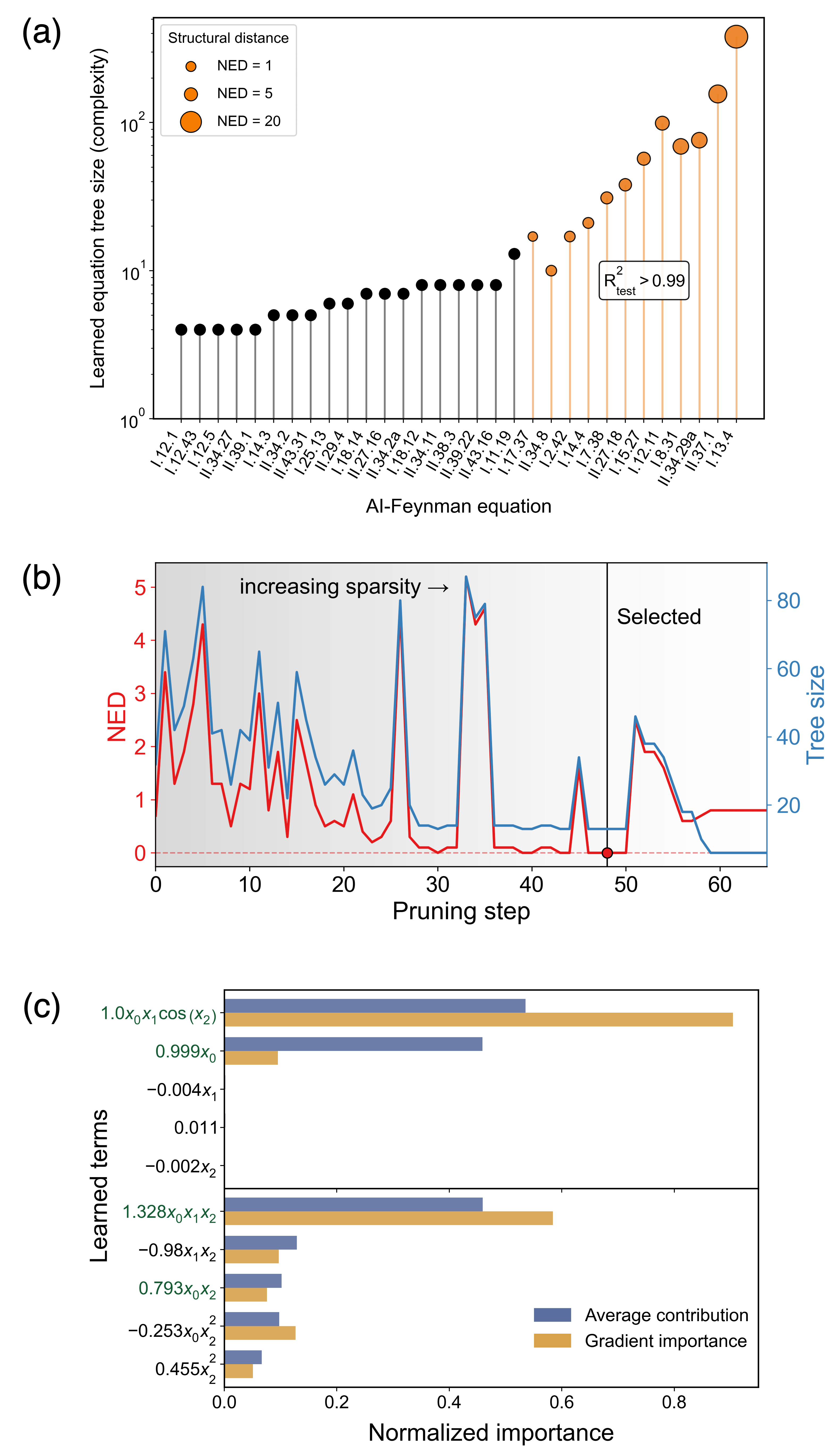}
  \caption{
  {\textbf{nn-EQL performance on a subset of the AI-Feynman benchmark.}}
  (a) Discovery of equations for 31 problems. Each point is the selected nn-EQL equation for one problem, plotted as normalized tree-edit distance (NED) to the ground truth versus learned equation complexity, as measured by tree size. All selected equations are accurate ($R^2_{\mathrm{test}} > 0.99$), and 19 of 31 exactly recover the ground-truth structure (NED $=0$). 
  (b) Candidate family generated by iterative pruning for equation I.11.19.
  (c) Term-level diagnostics for two selected equations: III.17.37 (top) and II.37.1 (bottom), where green terms are terms present in the ground truth equation.
  }
  \label{fig:ai_feynman_benchmark}
\end{figure}

We evaluate nn-EQL on a subset of the AI-Feynman benchmark to assess its ability to \textbf{generate} and \textbf{select} compact symbolic expressions.
This subset consists of 31 equations with varying levels of complexity.
Our first observation is that nn-EQL always discovers equations that are accurate, with $R^2_{\mathrm{test}} > 0.99$.
Out of the 31 equations tested, nn-EQL exactly recovers the ground truth structure for 19, corresponding to a normalized tree-edit distance (NED) of zero, see Fig.\ref{fig:ai_feynman_benchmark}(a).
The remaining equations span a broad range of structural distances and learned equation complexities.
This results shows that a highly accurate selected equation is not necessarily structurally equivalent to the ground truth equation.

Fig.\ref{fig:ai_feynman_benchmark}(b) illustrates how the candidate family is generated for one representative problem, equation I.11.19, $A = x_1y_1 + x_2y_2 + x_3y_3$ which represents a dot product between two vectors.
Iterative pruning progressively increases network sparsity (gray shade) and produces a sequence of symbolic expressions with different tree sizes and structural distances from the ground truth.
Neither quantity varies monotonically with pruning, emphasizing that pruning generates a family of competing hypotheses rather than a single trajectory toward the correct expression.
An accuracy-complexity model \textbf{selection} criterion can then be used to select a candidate from this family, although the choice depends on the relative penalty assigned to complexity.

However, Fig.\ref{fig:ai_feynman_benchmark}(a) does not indicate which components of a non-exact equation are scientifically meaningful.
For this, we \textbf{diagnose} candidate equations through the contribution and input-scaled sensitivity of their individual additive terms, Fig.\ref{fig:ai_feynman_benchmark}(c), for two example equations.
For equation III.17.37 $f = \beta(1+\alpha \rm cos(\theta))$ (the top panel), representing the angular distribution of particle decay products, the learned expression has a nonzero NED, yet the two terms present in the ground truth equation account for essentially 100\% of both measures, while the additional learned terms are negligible.
The diagnostics therefore indicate that the underlying relationship has been effectively recovered despite the small residual terms.
The conformal analysis further solidifies this, where we find that removing the two true structural terms increased the 90\% residual bound from \(0.018\) to \(12.0\) and \(4.7\), whereas removing small spurious linear terms reduced it.
This shows that conformal diagnostics can provide evidence for whether individual learned terms are genuinely supported by the data.
For equation II.37.1 $E_n = \mu(1+\chi)B$ (the bottom panel), representing the potential energy of a magnetic moment in a field, the selected equation contains both terms of the ground-truth equation, including a dominant $x_0x_1x_2$ contribution, but several additional terms also retain an appreciable contribution and input-scaled sensitivity.
Conformal analysis here shows that the dominant term $1.328 x_0x_1x_2$ has the highest $\Delta q$ (70.07), implying that removing this term drastically degrades the equation.
The other `correct' term $0.793 x_0 x_2$ has a more modest $\Delta q$ (11.91), comparable to the $x_1x_2$ and the $x_0x_2^2$ terms, implying that these terms are all equally important for the equation, and removing any one of them degrades the equation, although in a more limited manner than the dominant term.

the learned expression contains identifiable elements of the correct relationship, but the diagnostics provide weaker evidence that the complete expression should be interpreted as the underlying equation.

On simple benchmarks such as AI-Feynman, search-based methods such as PySR perform exceptionally well, even with noise in the dataset, and often recover the exact structure more reliably than nn-EQL.
Twelve of the 31 selected equations reach $R^2_{\text{test}} > 0.99$ without exactly recovering the ground-truth structure, and the NED varies non-monotonically along the pruning path. 
Therefore, one could ascribe the presence of additional terms discovered by nn-EQL, Fig. \ref{fig:ai_feynman_benchmark}(b), as an imperfection in our equation generation and selection process.
However, we show next that `imperfection' in equation generation/selection is only relevant for known benchmarks, and that for most scientific discovery tasks, equations with ambiguous terms allow us to rank competing mechanisms.

\subsection*{Learning equations from atomistic data of alloys}

We now consider a scientific setting in which the underlying equation is not known, and various ambiguous expressions can all fit the data well.
We learn relationships between grain-boundary energy and misorientation angle, $\theta$, from 96 atomistic Cu symmetric-tilt grain-boundary configurations spanning misorientation angles from 0 to 180$^\circ$.
Details of the atomistic grain-boundary simulations are provided in the Appendix.
For small misorientation angles, grain-boundary energy is described by the Read-Shockley relationship \cite{hirth1983theory}, whereas no single general functional form describes the full angular range considered here.

\begin{figure}[!ht]
\centering
  \includegraphics[width=\columnwidth]{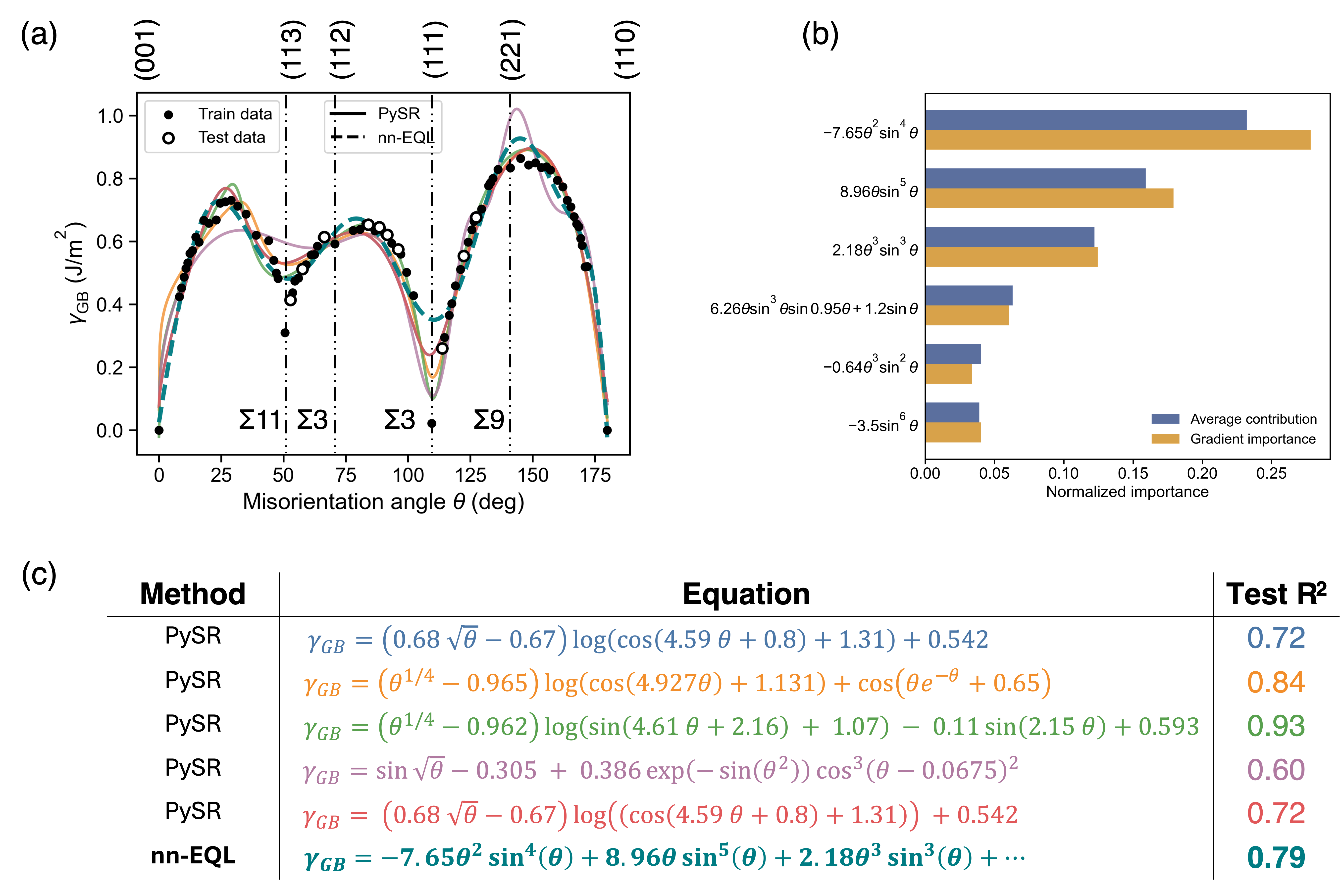}
  \caption{
  \textbf{nn-EQL performance on atomistic simulations of Cu $\langle 110 \rangle$ symmetric tilt grain boundaries.}
  (a) Grain-boundary energy $\gamma_{\mathrm{GB}}$ versus misorientation angle $\theta$ for 96 atomistic configurations, with curves from independent PySR runs and from nn-EQL overlaid on the data.
  (b) Term-level diagnostics for the nn-EQL expression.
  (c) Equations discovered using various search parameters for PySR, alongside the equation discovered with the nn-EQL (bold), along with $R^2$ on test set.
  }
  \label{fig:eql_gb}
\end{figure}

Figure~\ref{fig:eql_gb}(a) illustrates the ambiguity that arises when equations are learned directly from these data.
Independent PySR runs generate several distinct expressions on the same dataset, with held-out $R^2$ values ranging from approximately 0.60 to 0.93.
These surface-energy expressions contain qualitatively different combinations of square-root, logarithmic, and trigonometric functions.
nn-EQL produces another structurally distinct hypothesis with $R^2_{\mathrm{test}} = 0.79$, lying within the predictive range of the PySR candidates.
The same 96 configurations are therefore described by several equations with different functional forms and comparable accuracy.
Its dominant structure instead consists of polynomial-trigonometric interactions involving $\theta$ and $\sin\theta$ (note that the equation assumes $\theta$ in radians), with the full equation in the Appendix.
\begin{align}
\gamma^{\mathrm{PySR}}_{\mathrm{GB}}(\theta) &= (0.68\sqrt{\theta}-0.67)\log\!\Big[ \cos(4.59\theta + 0.8) + 1.31 \Big] + 0.54 \\[4pt]
\gamma^{\mathrm{nn\text{-}EQL}}_{\mathrm{GB}}(\theta) &= -7.65\,\theta^2\sin^4\theta + 8.96\theta\sin^5\theta + 2.18\theta^3\sin^3\theta \\ &+  \nonumber 6.26\theta\sin^3\theta \sin(0.95\theta+1.20\sin\theta) +\cdots
\end{align}

Thus, predictive accuracy alone does not determine a unique functional explanation of the observed grain-boundary energies, and the presence of competing terms is expected in scientific discovery settings.

Predictive accuracy cannot resolve this ambiguity, but term-wise diagnostics can narrow it.
We therefore examine the contributions, sensitivities, and conformal support of the terms in the nn-EQL equation.
Figure~\ref{fig:eql_gb}(b) ranks the leading terms according to their contribution and input-scaled sensitivity.
Although the expanded nn-EQL expression contains many terms, both diagnostics are concentrated in a substantially smaller subset.
The largest contributions arise from terms including
$-7.65\theta^2\sin^4\theta$,
$8.96\theta\sin^5\theta$, and
$2.18\theta^3\sin^3\theta$.
The agreement between contribution and gradient-based rankings provides evidence that these functional motifs are responsible for a substantial fraction of the learned response.

This example therefore illustrates the distinction between \textbf{selecting} an accurate equation and \textbf{diagnosing} what that equation implies.
Several structurally different expressions remain compatible with the available data.
In the nn-EQL equation, term-wise diagnostics single out three polynomial–trigonometric terms that carry most of the response, and these terms are relationships that can be tested in subsequent simulations or experiments.

\subsection*{Learning equations from high-dimensional data}\label{sec:nnEQL-highdim}

\subsubsection*{Example I: Mechanical MNIST dataset}

\begin{figure}[!ht]
\centering
  \includegraphics[width=0.9\columnwidth]{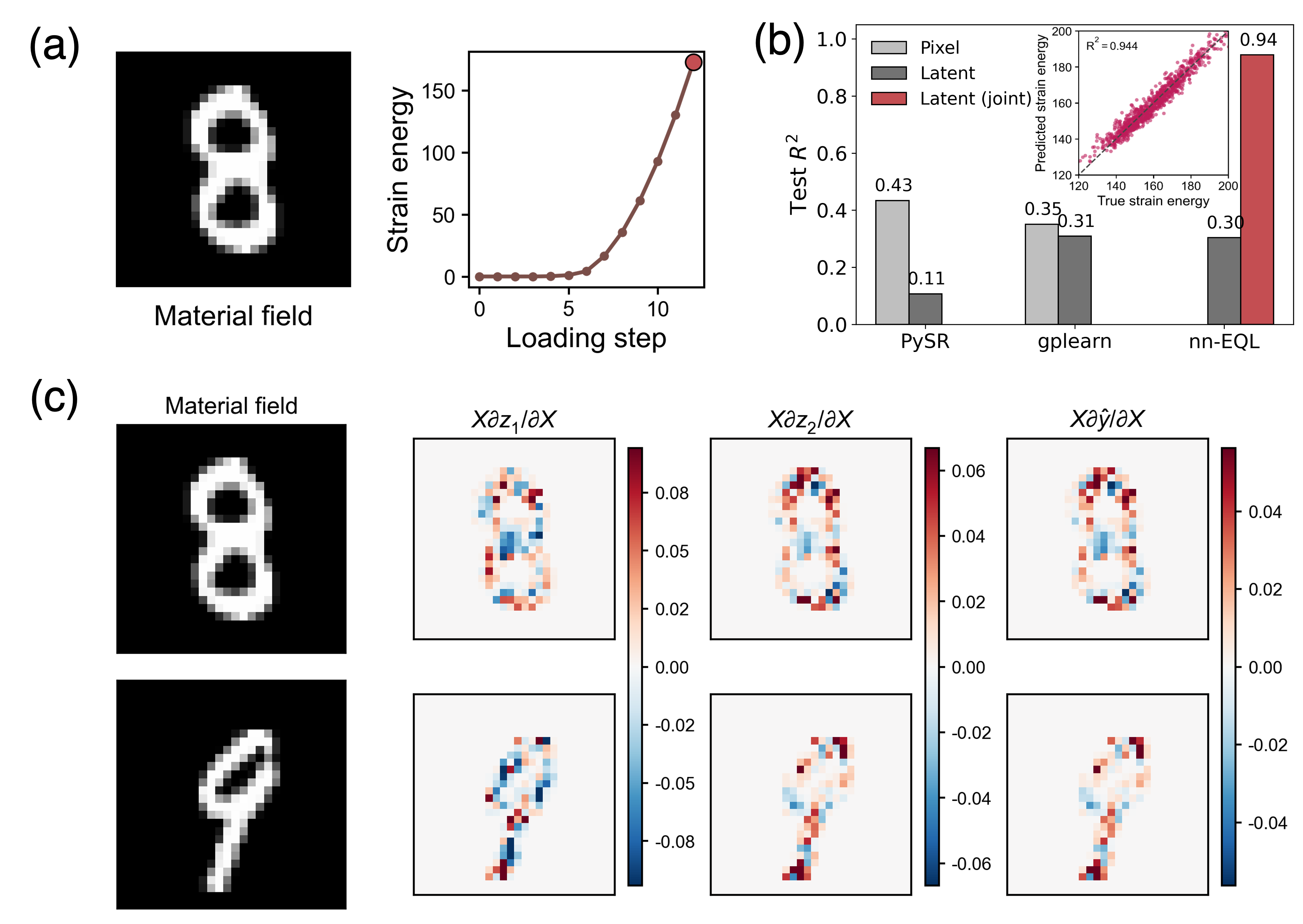}
  \caption{
  \textbf{Equation learning from high-dimensional Mechanical MNIST material fields.}
  (a) Example $28\times28$ material property field and corresponding strain energy response over the loading trajectory. The enlarged marker denotes the final loading step used as the prediction target.
  (b) Test $R^2$ for various equation learning methods, both in pixel space and in latent space. Inset shows predicted versus true final strain energy for the compact nn-EQL model, evaluated on 1\,000 held out test data points.
  (c) Signed input-scaled sensitivities for two representative material fields. Red and blue regions indicate pixels that locally increase or decrease the corresponding latent variable or predicted strain energy.
}
  \label{fig:eql_mmnist}
\end{figure}

Many materials datasets are not naturally represented by a small set of scalar descriptors.
Microstructure images, spectra, and other field data may contain hundreds or thousands of input dimensions, making direct equation search difficult and raising a more fundamental question: \emph{what variables should the equation be written in?}
We illustrate this challenge using the simple Mechanical MNIST dataset \cite{lejeune2020mechanical}, in which each sample is a $28\times28$ material-property field and the target is the strain energy at the final loading step.
An example material field and its strain energy trajectory are shown in Fig.~\ref{fig:eql_mmnist}(a).

A direct equation learner treats the 784 pixels as independent candidate variables $\hat{y} = f(x_1,\ldots,x_{784})$, and using PySR gives:
\begin{equation}
\hat{y}_{\mathrm{PySR,pixel}} = \sqrt{x_{329}} + \sqrt{x_{326}+x_{576}+x_{579}} + 141.93,
\end{equation}
with a test $R^2=0.434$.
Although compact, this expression selects a small number of individual pixels and does not encode the spatial organization of the material field.
This illustrates a limitation of applying equation learning, namely genetic programming, directly to high-dimensional structured data, where the equation learner must simultaneously identify relevant spatial information and construct the functional relationship.

A natural alternative is to first compress the image into a low-dimensional latent representation $\mathbf{z}$, and subsequently learn an equation in $\mathbf{z}$.
To this end, we trained an unsupervised autoencoder with a four dimensional latent space and subsequently applied PySR to the frozen latent variables.
Despite reconstruction errors of only $0.024$ and $0.032$ on the training and test sets, respectively, the resulting equation was
\begin{equation}
    \hat{y}_{\mathrm{AE+PySR}} = 161.56-z_3
\end{equation}
with test $R^2=0.107$.
A latent space that reconstructs the material field accurately does not need to be organized around the strain energy, so the variables of the equation must be learned with the prediction target in view.
We note that this result can be substantially improved if the representation $z$ learned by the autoencoder is conditioned by the prediction task.
A supervised convolutional encoder trained to predict the strain energy generated a four-dimensional latent representation in which a subsequent PySR search produced an equation with a test $R^2\simeq0.958$. 

Instead, with nn-EQL, we can integrate representation learning and equation learning into a single model,
\begin{equation}
    X \xrightarrow{g_\phi} \mathbf{z} \xrightarrow{f_{\mathrm{EQL}}} \hat{y},
\end{equation}
where a standard convolutional encoder is trained jointly with the nn-EQL network.
Note that we use a neural network based encoder, although our approach is flexible to many other dimensionality reduction techniques such as principal component analysis (PCA), and graph based autoencoders.
Following training of the dense nn-EQL, the encoder is frozen and the equation network is progressively pruned, producing a family of equations with increasing sparsity. A highly pruned member of this family contains only four terms,
\begin{equation}
\boxed{
\hat{y}_{\mathrm{nn\text{-}EQL}} = 1.213\,z_1^2 + 2.274\,z_1z_2 + 11.363\,z_2 +158.080}
\label{eq:mechanical_mnist}
\end{equation}
and achieves $R^2=0.944$ and RMSE $=3.64$ on 1000 held out images, as shown by the parity plot in Fig.~\ref{fig:eql_mmnist}(b).
We note that the poor performance of PySR on high-dimensional datasets also extends to other packages such as gplearn \cite{stephens_gplearn}, as indicated in Fig.~\ref{fig:eql_mmnist}(b).

Because the latent variables remain differentiable functions of the original image, \textbf{interpretation} can be propagated through the learned representation.
We therefore evaluate the input-scaled sensitivities
\begin{equation}
    X\odot\frac{\partial z_k}{\partial X}
    \qquad\mathrm{and}\qquad
    X\odot\frac{\partial \hat{y}}{\partial X},
\end{equation}
which identify image regions that locally increase or decrease each latent coordinate and the final prediction. Figure~\ref{fig:eql_mmnist}(c) shows these quantities for two geometrically distinct material fields.
The sensitivity maps show different spatial roles for $z_1$ and $z_2$, while the output sensitivity combines these latent responses according to Eq.~\eqref{eq:mechanical_mnist}.
This provides a route from a compact equation in latent space back to the original high-dimensional material structure.

Term-level diagnostics, as done for earlier cases, provide complementary information on the learned equation.
Among the non-constant terms in Eq.~\eqref{eq:mechanical_mnist}, the linear $z_2$ term accounts for approximately $81.0\%$ of the normalized input-scaled sensitivity, followed by the $z_1z_2$ interaction at $14.8\%$ and the $z_1^2$ term at $4.2\%$.
The $z_2$ term is also strongly correlated with the target ($r=0.969$), suggesting that $z_2$ acts as the dominant mechanical coordinate, while $z_1$ primarily modifies the prediction through a weaker nonlinear correction and interaction.

\subsubsection*{Example II: Learning hardness from microstructural features in nanocrystalline alloys}

The previous example is a favorable setting for equation learning from high-dimensional data, with noise-free spatial fields and a large training dataset with known, smooth constitutive laws solved by finite element simulations.
In contrast, experimentally obtained materials datasets rarely share these properties.
We therefore apply nn-EQL to an experimental dataset of nanocrystalline Pt-Au thin films.
This dataset is noisy (X-ray diffraction (XRD) for microstructure, nanoindentation for hardness) and sparse (784 films), and the input is a one-dimensional spectrum traditionally interpreted through established human-identified descriptors (peak positions, widths, and intensity ratios).
Most importantly, no governing equation links diffraction patterns to hardness, which instead results from simultaneous Hall-Petch strengthening, residual stress, texture, and solid-solution effects.
Details on this dataset are provided in the Methods section, and in previous work
\cite{desai2024navigating, adams2024guided}.

\begin{figure}[!ht]
\centering
  \includegraphics[width=0.81\textwidth]{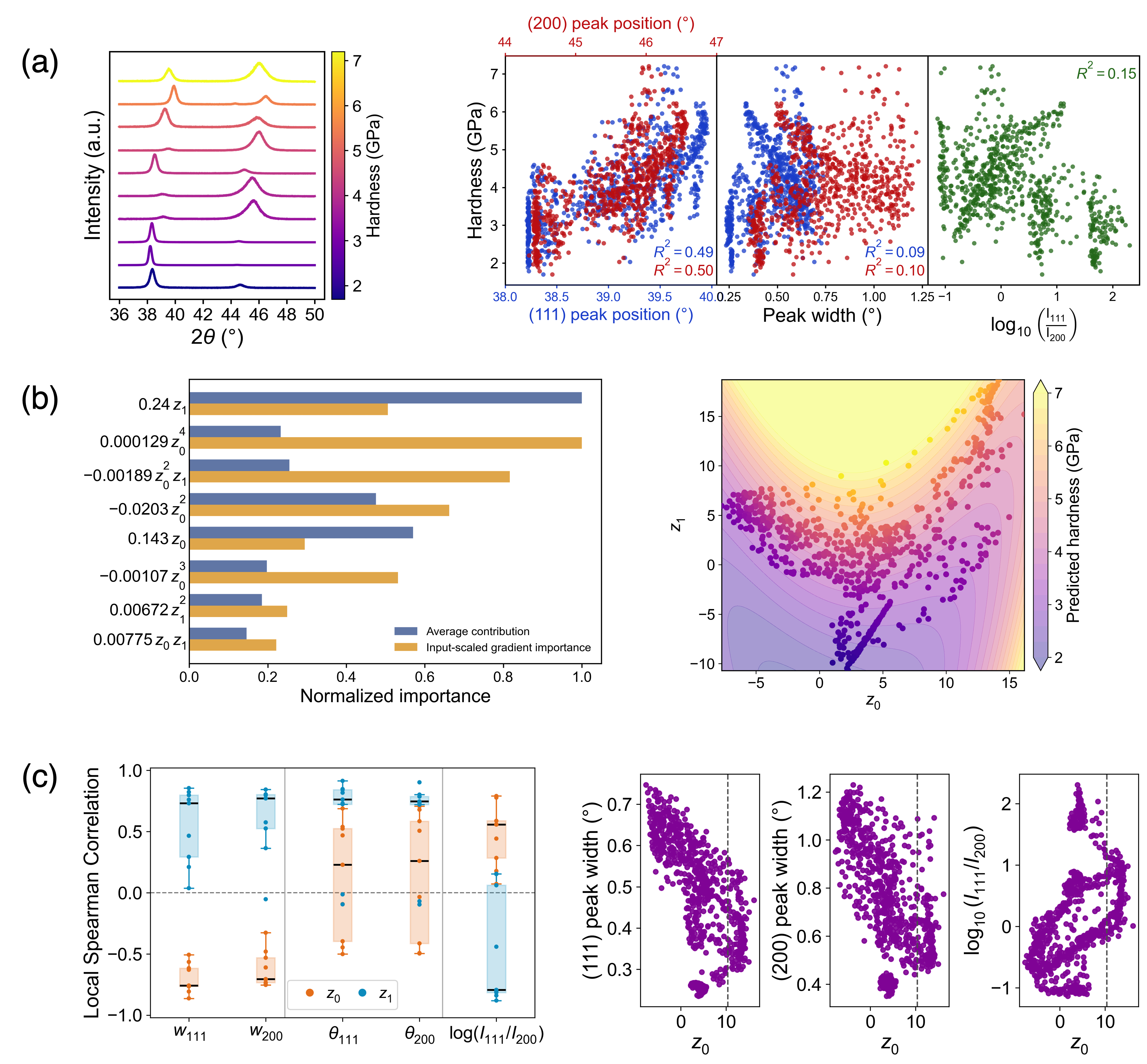}
  \caption{
  \textbf{Learning equations from experimental data relating X-ray diffractograms to hardness for nanocrystalline binary alloys.}
  (a) (Left) Sample X-ray diffractograms ordered by increasing hardness. (Right) Correlations between hardness and human-identified key features of diffractograms, including peak locations, peak widths, and intensity ratio. 
  (b) (Left) Term-wise contributions and sensitivities, highlighting dominant terms governing hardness. (Right) Hardness as a function of $z_0, z_1$, as learned by nn-EQL (heatmap, see Eq\@.~\eqref{eq:xrd_hardness}) with experimental data overlaid (data points).
  (c) (Left) Interpreting the linear terms in the learned equation with local spearman correlations between latent variables and human-identified features. Spearman correlations are calculated locally by holding one variable (e.g. $z_0$) nearly constant, and varying $z_1$. Correlations are evaluated at 10 percentile levels, establishing the spread of correlations in various hardness regimes. (Right) Interpreting microstructural features where the $\rm 4^{th}$ order term dominates the linear term in the learned equation.
}
  \label{fig:eql_xrd}
\end{figure}

Figure \ref{fig:eql_xrd}(a), left panel, shows sample X-ray diffractograms for 10 samples, ordered and colored by hardness.
The X-ray diffractograms are characterized by two dominant peaks, corresponding to the (111) and the (200) planes, at $2\theta \approx 38^\circ$ and $46^\circ$, respectively.
Traditional human-identified features of these patterns include peak locations $\theta_{(111)}, \theta_{(200)}$ (indicative of tensile or compressive strain), peak widths $w_{(111)}, w_{(200)}$ (indicative of effective crystallite size), and intensity ratio $\log(I_{(111)}/I_{(200)})$  (denoting film texture).
Figure~\ref{fig:eql_xrd}(a), right panel, shows that no single descriptor predicts hardness well, with the highest $R^2$ value being 0.49 and 0.50, for the (111) and (200) peak locations respectively, while peak widths and the intensity ratio reach only $R^2 = 0.09$-$0.15$.
The lack of strong individual correlations indicate that hardness therefore depends on a combination of these descriptors.
We note that X-ray diffractograms are one modality to capture microstructural features, the full dataset includes X-ray reflectivity as a surrogate for density \cite{adams2024guided, addamane2026high}.
A complete description of properties such as hardness would require models that ingest multimodal datasets.
Regardless, black-box machine learning techniques have helped predict properties from microstructural information \cite{vizoso2023connecting,desai2024navigating,vizoso2025decoding}, but they lack interpretability without post-hoc sensitivity techniques.
Here, we investigate whether nn-EQL can learn an equation for hardness directly from the full X-ray diffractograms, whether the data supports its terms, and whether its latent variables correspond to the human-identified features.
As with the mechanical MNIST example, we trained an encoder jointly with nn-EQL, here compressing each diffractogram into a two-dimensional latent space $\mathbf{z} = (z_0, z_1)$.
Details on the dataset creation, and the encoder model, are provided in the Methods section.

We find that the nn-EQL learns the following equation describing hardness $\hat{H}_{\mathrm{nn\text{-}EQL}}$:
\begin{align}
\boxed{
\begin{aligned}
\hat{H}_{\mathrm{nn\text{-}EQL}} 
&= 0.24z_1 + 0.143z_0 + 0.00013z_0^4 - 0.0011z_0^3\\& - 0.0019z_0^2 z_1 - 0.0203z_0^2 + 0.0078z_0z_1 + 0.0067z_1^2 + 3.9
\end{aligned}
}
\label{eq:xrd_hardness}
\end{align}
which fits the training set ($R^2 = $ 0.87) and generalizes to the validation ($R^2 = $ 0.66) and test sets ($R^2 = $ 0.73).
The term-wise diagnostics (Fig.~\ref{fig:eql_xrd}(b), left panel) show that the linear terms $0.24z_1$ and $0.143z_0$ dominate the contributions to hardness, while nonlinear terms, in particular $0.00013z_0^4$, dominate the sensitivities.
The response surface learned by the equation is visualized in Fig.~\ref{fig:eql_xrd}(b), right panel.
We find that higher hardness regions are generally found with increasing $z_1$, with a nearly quadratic (in $z_0$) contour, supporting the linear term in $z_1$ and the quadratic term in $z_0$ in the equation learned by nn-EQL.
Interestingly, the microstructural features in some films occupy a linear manifold in the latent space, corresponding to narrow range of alloy compositions.

To further interpret the learned equation, we inspect the dominant terms in the equation, see Fig. \ref{fig:eql_xrd}(c).
We first compute Spearman correlations between the latent variables learned by the encoder $z_0, z_1$ and the human-identified features $w_{(111)}, w_{(200)}, \theta_{(111)}, \theta_{(200)}, \rm log (\frac{I_{(111)}}{I_{(200)}})$, denoting the full-width half-maximum (FWHM) widths of the (111) and (200) peaks, the peak locations, and the intensity ratio.
The 784 films span a wide range of processing conditions, so several microstructure-hardness regimes may coexist, with correlations that are strong in one regime and weak in another.
We instead exploit the two-dimensional latent space and compute local correlations.
We fix $z_0$ at a given percentile and vary $z_1$.
Because the data is sparse, a smooth path in latent space would pass through points with no measured diffractogram.
We therefore select, at each point along the path, the nearest measured film in latent space and use its descriptors.
Repeating this at 10 percentiles of $z_0$ yields 10 local Spearman correlations, whose spread appears in Fig.~\ref{fig:eql_xrd}(c), left panel.

We find that at fixed $z_0$, increasing $z_1$ broadens both peaks at every $z_0$ percentile.
From the Scherrer equation, broader peaks indicate smaller effective crystallite sizes, and from the Hall-Petch relationship, higher hardness, consistent with the dominant term $0.24z_1$ in our equation.
Interestingly, increasing $z_1$ also shifts both peaks to higher angles, indicating smaller interplanar spacing and hence compressive strain, consistently across all $z_0$ percentiles.
Finally, increasing $z_1$ lowers the intensity ratio, i.e., strengthens the (200) texture.
The term $0.24z_1$ therefore encodes three aspects of higher hardness: smaller crystallites, as expected from Hall-Petch, and two less conventional ones, compressive residual strain and (200) texture.

At fixed $z_1$, increasing $z_0$ narrows both peaks (orange in Fig.~\ref{fig:eql_xrd}(c), left panel), correlates weakly with peak positions, and raises the intensity ratio.
That is, $z_0$, and by extension, the linear term $0.143 z_0$, tell us that hardness can increase from reduced peak widths (higher effective crystallite sizes), but mostly when the films have strong (111) texture.

The hardness equation Eq. \eqref{eq:xrd_hardness} not only has linear contributions, but it also contains nonlinear and coupled terms.
We focus on the quartic term $0.00013z_0^4$, the strongest nonlinear term in the diagnostics.
Compared to the linear term $0.143 z_0$, this 4th order term dominates at $z_0 > 10.3$, which corresponds to specific microstructural regimes, see the right panel of Fig. \ref{fig:eql_xrd}(c).
Films in this regime have strong (200) texture ($\log(I_{(111)}/I_{(200)}) < 1$), (111) peak widths generally below $0.5^\circ$, and (200) peak widths generally below $0.8^\circ$.
Through the Scherrer equation, these widths correspond to effective crystallite sizes above $\sim$17~nm for (111) and $\sim$11~nm for (200).

nn-EQL thus learns an explicit equation for hardness from X-ray diffractograms, and its latent variables map onto peak widths, peak positions, and the intensity ratio.
The discovered equation recovers the expected link between small crystallites (broad peaks) and high hardness, but also associates compressive strain (peaks shifted to higher angles) and (200) texture with higher hardness in Pt-Au.
Finally, the learned equation isolates a regime with relatively strong texture (200) and crystallite sizes greater than 11-17 nm, in which the fourth-order term in $z_0$ governs the hardness response.
Each of these relationships can be tested directly through targeted deposition and characterization.

\section*{Discussion}
This work presents nn-EQL as an equation learning workflow to learn equations from high-dimensional materials datasets.
The nn-EQL workflow constitutes equation \textbf{generation} via flexible neural networks, and equation \textbf{selection} via a heuristic minimum-description-length (MDL) principle that balances accuracy and complexity.
Going beyond traditional symbolic regression frameworks, nn-EQL assists scientific discovery by providing \textbf{diagnosis} and \textbf{interpretation} for the learned equations, using term-wise contributions, input-scaled sensitivities, and uncertainties.
We demonstrate nn-EQL on traditional benchmark data as well as high-dimensional materials simulations and experimental datasets.
Our results on the mechanical MNIST and Pt-Au case studies above demonstrate that the nn-EQL can be extended beyond scalar descriptors by mapping high-dimensional data into a compact latent space and then learning equations over learned representations.
This same strategy points toward a broader class of materials data commonly found in workflows for scientific discovery~\cite{desai2024navigating}. For instance, imaging modalities such as transmission or scanning electron microscopy provide spatially resolved structural and defect information, while spectroscopic time series such as electron energy loss or Raman spectroscopy provide dynamic signatures that are difficult to distill in a handful of scalar values.
Coupling representation learning and data fusion~\cite{trask2024unsupervised, walker2024unsupervised} with nn-EQL offers a route to discover compact equations directly from these modalities, provided that their latent representation can be constructed or constrained so that it remains physically interpretable as demonstrated by our Pt-Au case study. 

A further extension concerns data of varying fidelity: atomistic, mesoscale, and continuum simulations, as well as experiments, often describe the same phenomenon at different resolutions and computational cost.
Multi-fidelity machine-learning concepts~\cite{liu2019multi,batra2020machine} could provide a basis for devising multi-fidelity equation learning strategies in which inexpensive, abundant low-fidelity data (e.g., coarse-grained or reduced-order simulations) are used to constrain the space of plausible functional forms, whereas sparse high-fidelity data, including ab-initio calculations, high-resolution imaging, and targeted experiments, discriminate among and refine the remaining hypotheses.
Combined with the diagnostic tools discussed above, such an approach would allow equation learners to be deployed hierarchically across scales and data sources, systematically identifying which structural, spectral, or multi-fidelity features are mechanistically relevant, and directing new simulations or experiments to resolve the ambiguity that remains.

Techniques such as active learning or Bayesian optimization can be coupled to equation learning to adaptively improve and refine equations with additional data.
Prior work on active symbolic regression \cite{ansari2022iterative, medina2023active} that focuses on strategies such as query by committee to learn equations from simulations.
Extending this further, the equation search process itself can be governed by Bayesian techniques \cite{bomarito2025bayesian}.
Bayesian techniques offer rigorous uncertainties to parameters and terms in equations, but their scalability to high-dimensional equation learning remains an open question.

Another promising direction is equation learning using large language models (LLMs), with recent work using LLMs to generate candidate equations as Python scripts, evaluating them on datasets, and using experience management to iterate over scripts \cite{shojaee_llm-sr_2025, shojaee2025llm-srbench}.
LLMs have the promise of connecting the equation discovery to prior scientific hypotheses in literature, translating conventional hypotheses from a natural language representation, to a numerical or programmable description that can be iterated over via search algorithms.
Similarly, agentic workflows have also shown the ability to recover benchmark equations accurately \cite{xia2026sr}, but extending these workflows beyond benchmarks and towards real-world domain applications is an open question.
Self-driving laboratories could use equations with term-wise uncertainties as explicit hypotheses, choosing the next experiment to test the term the data supports least.

\section*{Materials and Methods}

\subsection*{Using the nn-EQL package}
\label{sec:nneql_code_walkthrough}
The nn-EQL codebase is available in a public repository on GitHub at~\url{https://github.com/mirage-scidac/nn-EquationLearner}.
Each nn-EQL experiment is specified by a YAML configuration file that defines the dataset, equation-learning architecture, optimization parameters, pruning schedule, and optional evaluation settings.
This separates the equation-search space from the training code and allows the same workflow to be applied across benchmark and materials datasets.
A minimal configuration has three primary components: \textit{data}, \textit{model}, and \textit{training}.

\begin{lstlisting}[style=nneql,language=yaml,caption={Example nn-EQL configuration.}]
# Data
data_path: "path/to/data"
data_cols: [0, 1, 2, 3]
train_samples: 1000
# Model
model:
  act_func: ['linear', 'sin', 'prod', 'linear']
  act_poly: ['linear', 'square', 'cube', 'linear']
  n_rep: 2
# Training and pruning
pretrain_epochs: 5001
learning_rate: 0.01
batch_size: 100
retrain_epochs: 500
prune_percentage: [0.1, 0.1]
target_sparsity: [0.9, 0.9]
# Evaluation
evaluation:
  expand_expression: true
  coefficient_tolerance: 0.001
\end{lstlisting}

The entries \texttt{act\_func} and \texttt{act\_poly} define the candidate mathematical operators available in the two equation learning layers.
Available operators include linear, product, polynomial, trigonometric, exponential, logarithmic, and related transformations. 
The parameter \texttt{n\_rep} controls the number of repetitions of the specified operator set and therefore the width of the search space.
Problem-specific preprocessing can additionally be specified in the configuration, e.g. introducing inverse or unary-transformed input features.
Training is launched from the command line as:
\begin{lstlisting}[style=nneql,language=bash]
python train.py --config configs/example.yml --seed 0 --work_dir work_dir/
\end{lstlisting}
After training, candidate equations may subsequently be ranked using a complexity-penalized selection score, evaluated using:
\begin{lstlisting}[style=nneql,language=bash]
python model_selection.py --work_dir work_dir/ --lambda_value 1.0
\end{lstlisting}
Selected models can be evaluated using
\begin{lstlisting}[style=nneql,language=bash]
python evaluate_checkpoint.py --config configs/example.yml --work_dir work_dir/ --checkpoint work_dir/models/_eql_model_pruning_step_$STEP.pth
\end{lstlisting}
Finally, diagnostics such as term-wise contributions and input-scaled sensitivities, as well as uncertainties, can be obtained by:
\begin{lstlisting}[style=nneql,language=bash]
python symbolic_diagnostics.py --config configs/example.yml --work_dir work_dir/evaluations/step_$STEP
\end{lstlisting}

\subsection*{Atomistic simulations of grain boundaries in copper}

Symmetric tilt grain boundaries (STGBs) in Cu were generated for the $[1\bar{1}0]$ tilt axis using a bicrystal construction, following the procedure of Tschopp and co-workers~\cite{tschopp2007structure,tschopp2015symmetric}.
Two half-crystals were rotated by $\pm\theta/2$ about the tilt axis, so that the boundary plane is a mirror plane of the bicrystal, with $\theta = 0^\circ$ corresponding to the perfect crystal (reference direction $[001]$).
A total of 96 boundaries were generated, spanning misorientation angles $\theta$ from $0^\circ$ to $180^\circ$.
The in-plane periodic lengths of each bicrystal were set by the two-dimensional periodicity of the coincidence-site lattice for the given $\theta$.
Periodic boundary conditions were applied in all three directions, so each cell contains two equivalent boundaries, and the half-crystal thickness was chosen large enough that the two boundaries do not interact.
Atomic interactions were described by the embedded-atom-method potential of Mishin for Cu~\cite{mishin2001structural}, which reproduces the lattice parameter, cohesive energy, elastic constants and stacking-fault energies.
Simulations were performed with the Large-scale Atomic/Molecular Massively Parallel Simulator (LAMMPS)~\cite{thompson2022lammps}.

The minimum-energy structure of each boundary was located with the $\gamma$-surface method~\cite{tschopp2007structure,tschopp2015symmetric}.
The upper grain was rigidly translated relative to the lower grain over a grid spanning the in-plane cell of the displacement-shift-complete (DSC) lattice, since translations by DSC vectors preserve the bicrystal periodicity.
At each grid point the atoms were relaxed along the boundary normal only, and the resulting energy defined the $\gamma$-surface.
The lowest-energy translations were then used as starting configurations for full relaxation, with all atoms free to move and the cell dimension normal to the boundary allowed to relax, using conjugate-gradient energy minimization at 0~K.

The grain-boundary energy was obtained as the excess energy relative to the bulk lattice,
\begin{equation}
\gamma_{\mathrm{GB}}(\theta) = \frac{\sum_{i=1}^{N}\left(e_i - e_{\mathrm{bulk}}\right)}{2A},
\end{equation}
where $e_i$ is the energy of atom $i$, $e_{\mathrm{bulk}}$ is the per-atom energy of perfect FCC Cu, $N$ is the number of atoms in the cell, and $A$ is the boundary area.
The factor of two accounts for the two boundaries in the periodic cell.
To cross-check our method, we confirmed that the $\Sigma3\,(111)$ coherent twin at $\theta=109.5^\circ$ and the $\Sigma11\,(113)$ boundary at $\theta=50.5^\circ$ appear as cusps in the resulting $\gamma_{\mathrm{GB}}(\theta)$ curve, consistent with prior work~\cite{tschopp2007structure,tschopp2015symmetric}.
The resulting pairs $(\theta,\gamma_{\mathrm{GB}})$ form the dataset used for equation learning.
The config file and training details for the equation learner can be found in the Github.

\subsection*{Finite-element simulations of a heterogeneous material}

The mechanical MNIST benchmark dataset represents a high-dimensional materials modeling dataset, in this case using finite element simulations.
Building on the well-known MNIST dataset, mechanical MNIST assumes that a handwritten digit image corresponds to a domain with two neo-Hookean hyperelastic materials.
The black background is assumed to be a soft material with Young's modulus $E = 1$, while the bright digit is assumed to be stiff with $E = 100$. The Poisson's ratio is fixed at 0.3.
Assuming a compressible neo-Hookean material, one can write strain energy
\begin{equation}
\psi = \frac{\mu}{2} \left[ F:F - 3 - 2 \rm ln(\rm det F) \right ] + \frac{\lambda}{2} \left[ \frac{1}{2}((\rm det F)^2 - 1) - \rm ln(\rm det F) \right ]
\end{equation}
where $\psi$ is the strain energy, $F$ is the deformation gradient, and $\mu, \lambda$ are the Lame parameters, related to the Young's modulus and Poisson's ratio as $E = \frac{\mu(3\lambda + 2\mu)}{\lambda + \mu}$ and $\nu = \frac{\lambda}{2(\lambda + \mu)}$.

Mechanical MNIST simulates the deformation in this heterogeneous material, assuming the bottom part of the simulation domain is fixed and studying uniaxial, biaxial, and other loadings. In this work we apply nn-EQL to the uniaxial loading dataset.
We predict the strain energy at 50\% displacement, i.e., a displacement of 14 compared to the original `material' length of 28.
Together, the pair of handwritten digit $X$ and strain energy at displacement 14 $y$ form the dataset $(X, y)$ used by the nn-EQL.
We refer the reader to \cite{lejeune2020mechanical} for additional details on the dataset and the simulations.
The config file and training details for the equation learner are found in the Github.

\subsection*{Combinatorial Pt-Au thin-film dataset}

The Pt-Au thin-film dataset used in this work was created by Adams et al. \cite{adams2024guided}, using high-throughput physical vapor deposition (PVD) to deposit 784 thin films across 7 wafers, with the goal of identifying Pt-Au alloys with high hardness (above 6~GPa) and low electrical resistivity, as candidates for next-generation interconnects and electrical contacts.
Deposition was conducted with the Kurt J. Lesker Inc. PVD-200 deposition system, and varying the Pt and Au sputter gun powers and angles resulted in films with varying composition and microstructure across the films.
Even for pure Pt and pure Au wafers, 112 films with varying process conditions (sputter gun voltages and background Ar gas pressure) were deposited to obtain a rich process-structure-property map.
Process conditions were further quantified using the Monte Carlo-based program SIMTRA \cite{depla2012magnetron}, which simulated the deposition of Pt/Au on the film substrate and atomic-level collisions between Pt/Au and the background Ar gas.
Structural fingerprints obtained for each film included X-ray diffractograms to measure crystal structure and texture, Wavelength Dispersive Spectroscopy (WDS) to measure composition, AFM to measure surface roughness, laser profilometry to measure thickness, and X-ray reflectivity (XRR) to measure density.
We note that XRR correlates strongly with density as studied by Addamane et al. \cite{addamane2026high}.
Properties measured included hardness, established by nanoindentation, and resistivity via four-point probe sheet resistance.
The config file and training details for the equation learner are found in the Github.

\subsection*{Details on encoder models}

Here we provide architecture and training details on encoder models used in the mechanical MNIST and the X-ray diffractogram examples.
The mechanical MNIST example used an encoder with two 2D convolutional layers, both with kernel size 3 and 8, 16 channels respectively, followed by ReLU activations and a pooling operation.
For the X-ray diffractogram example, the encoder model used four 1D convolutional layers, with kernel size 11 and padding 5, each layer followed by a ReLU activation function and a 1D pooling operation.
The number of channels (1 in the input data) varied as 8, 16, 16, and 16 respectively.
In both encoder models, a linear head follows the convolution and pooling operations, consisting of two linear layers with a ReLU activation function were used to learn the latent space.
The final linear layer did not have an activation function, giving the latent space unrestricted range.

\clearpage 

%
\bibliography{rsc} 
\bibliographystyle{sciencemag}
%


\section*{Acknowledgments}

The authors would like to thank Prof\@.~Krishna Garikipati from the University of Southern California for constructive feedback and discussion during the preparation of this work.

\paragraph*{Funding:}
All the authors in this work are supported by the U.S\@.~Department of Energy, Office of Science, Office of Advanced Scientific Computing Research and  Office of Basic Energy Sciences, Scientific Discovery through Advanced Computing (SciDAC) program under the MIRAGE project.
The atomistic and Pt-Au thin-film datasets used in this work were generated through the Center for Integrated Nanotechnologies (CINT) program, an Office of Science user facility operated for the U.S\@.~Department of Energy.
This article has been authored by an employee of National Technology \& Engineering Solutions of Sandia, LLC under Contract No\@.~DE-NA0003525 with the U.S\@.~Department of Energy (DOE).
The employee owns all right, title, and interest in and to the article and is solely responsible for its contents. The United States Government retains and the publisher, by accepting the article for publication, acknowledges that the United States Government retains a non-exclusive, paid-up, irrevocable, world-wide license to publish or reproduce the published form of this article or allow others to do so, for United States Government purposes.
The DOE will provide public access to these results of federally sponsored research in accordance with the DOE Public Access Plan {https://www.energy.gov/downloads/doepublic-access-plan.}

\paragraph*{Author contributions:}
S.D., P.P.I, and R.D. conceived the idea; R.D acquired the funding for this work; P.J performed the AI-Feynman benchmarking and visualized the results; S.D developed the codebase and some visualizations. All authors contributed to the discussion and writing of the paper.

\paragraph*{Competing interests:}
There are no competing interests to declare.
\paragraph*{Data and materials availability:}
The code and data used to support the findings illustrated in this article are available in the public repository on GitHub at~\url{https://github.com/mirage-scidac/nn-EquationLearner}.

\end{document}